\documentclass{article}

\usepackage[preprint]{neurips_2026}

\usepackage[utf8]{inputenc} 
\usepackage[T1]{fontenc}    
\usepackage{hyperref}       
\usepackage{url}            
\usepackage{booktabs}       
\usepackage{amsfonts}       
\usepackage{nicefrac}       
\usepackage{microtype}      
\usepackage[table]{xcolor}

\usepackage{amsmath}
\usepackage{multirow}
\usepackage{colortbl}
\usepackage{pifont}

\usepackage{graphicx}

\usepackage{fontawesome5}

\hypersetup{
    colorlinks=true,
    citecolor=green
}

\title{Beyond the Last Layer: Multi-Layer Representation Fusion for Visual Tokenization}

\author{%
  Xuanyu Zhu\textsuperscript{1,2}\thanks{Equal contribution.} \quad  
  Yan Bai\textsuperscript{2}\footnotemark[1] \quad
  Yang Shi\textsuperscript{1,}\thanks{Project Leader}  \quad 
  Yihang Lou\textsuperscript{1} \quad
  \\
  \textbf{  
  Yuanxing Zhang\textsuperscript{1} \quad
  Jing Jin\textsuperscript{3} \quad 
  Yuan Zhou\textsuperscript{4,}\thanks{Corresponding author} \quad
  } \\[0.2cm]
  \textsuperscript{1}Peking University \quad
  \textsuperscript{2}Meituan Inc \quad
  \textsuperscript{3}Tsinghua University \quad
  \textsuperscript{4}IGDL 
  \\[0.2cm]
    {\centering}
    \url{https://github.com/zhuzil/DRoRAE}
}

\begin{document}

\maketitle

\begin{abstract}
Representation autoencoders that reuse frozen pretrained vision encoders as visual tokenizers have achieved strong reconstruction and generation quality. However, existing methods universally extract features from only the last encoder layer, discarding the rich hierarchical information distributed across intermediate layers. We show that low-level visual details survive in the last layer merely as attenuated residuals after multiple layers of semantic abstraction, and that explicitly fusing multi-layer features can substantially recover this lost information. We propose \textbf{DRoRAE} (\textbf{D}epth-\textbf{Ro}uted \textbf{R}epresentation \textbf{A}uto\textbf{E}ncoder), a lightweight fusion module that adaptively aggregates all encoder layers via energy-constrained routing and incremental correction, producing an enriched latent compatible with a frozen pretrained decoder. A three-phase decoupled training strategy first learns the fusion under the implicit distributional constraint of the frozen decoder, then fine-tunes the decoder to fully exploit the enriched representation. On ImageNet-256, DRoRAE reduces rFID from 0.57 to 0.29 and improves generation FID from 1.74 to 1.65 (with AutoGuidance), with gains also transferring to text-to-image synthesis. Furthermore, we uncover a log-linear scaling law ($R^2{=}0.86$) between fusion capacity and reconstruction quality, identifying \textit{representation richness} as a new, predictably scalable dimension for visual tokenizers analogous to vocabulary size in NLP.
\end{abstract}

\section{Introduction}

The image tokenizer maps pixels into a compact latent space and defines the quality ceiling of modern visual generation systems~\cite{rombach2022ldm, peebles2023dit}. 
A recent line of work~\cite{yu2025repa, yao2025vavae, zheng2025rae, gong2026rpiae} has demonstrated that leveraging pretrained vision foundation models (VFMs) such as DINOv2~\cite{oquab2024dinov2} as the tokenizer's latent space yields substantial improvements in both reconstruction fidelity and downstream generation quality over conventional learned tokenizers trained from scratch.

\begin{figure}[t]
    \centering
    \includegraphics[width=\linewidth]{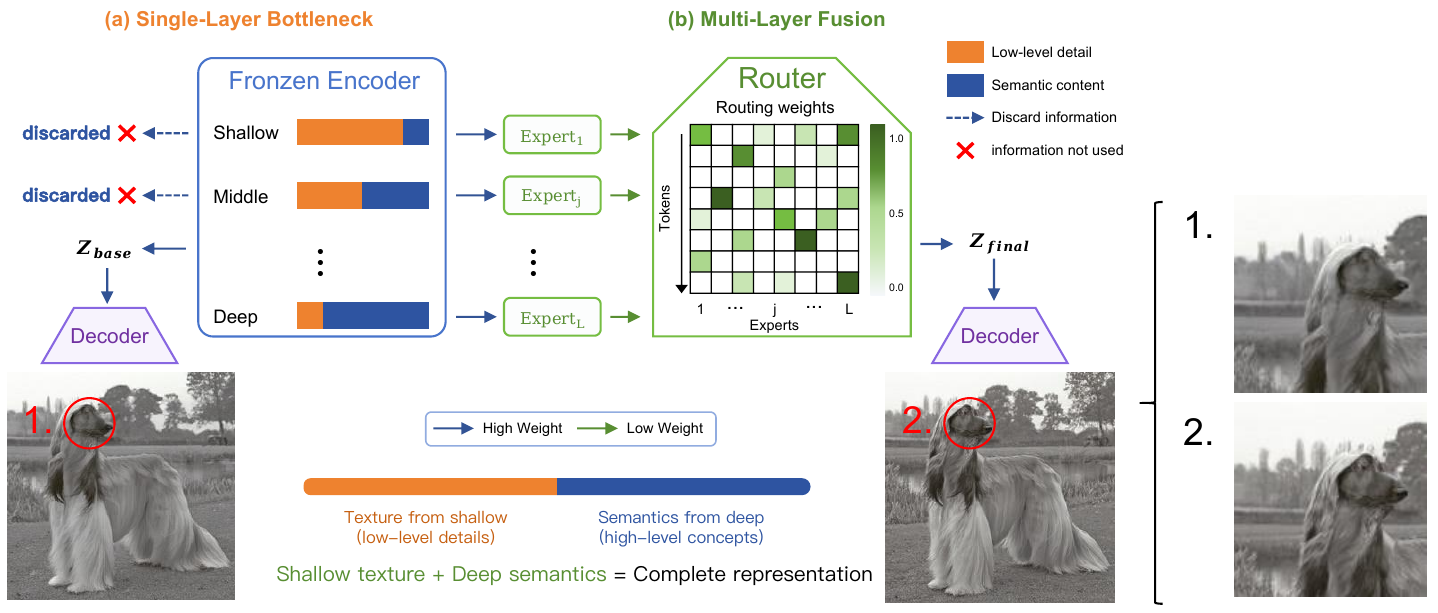}
    \caption{\textbf{Motivation.} Existing representation autoencoders extract only last-layer features, where low-level details are progressively diluted by semantic transformations. DRoRAE fuses features across all layers to assemble a richer latent per spatial token.}
    \label{fig:motivation}
    \vspace{-0.5cm}
\end{figure}

Despite their success, all existing VFM-based tokenizers share a common design choice: they extract features exclusively from the \textit{last layer} of the encoder. While this is the natural output of any vision model, last-layer features are primarily optimized for high-level semantics rather than low-level visual details such as textures, edges, and color gradients. Recent analysis~\cite{longcatnext2025} reveals that low-level information survives in the last layer only as a structural consequence of residual connections, a passive pathway that becomes increasingly lossy as each successive layer superimposes semantic transformations onto the residual stream. Shallower layers, by contrast, retain this information with far greater fidelity (Figure~\ref{fig:motivation}), yet single-layer tokenizers discard it entirely.

This observation suggests a natural direction: explicitly fusing features from multiple depth levels to assemble a latent representation richer than any single layer can provide. Moreover, multi-layer fusion introduces two quantifiable capacity axes, the number of fused layers and the per-layer expert capacity, which together define the \textit{representation richness} of the tokenizer. An analogous concept has been explored for NLP tokenizers~\cite{huang2025overtokenized}, where increasing the input vocabulary size (representation richness in the text domain) yields predictable, log-linear improvements in downstream loss. Whether such a scaling law also exists for visual tokenizers remains an open question.

Realizing multi-layer fusion in practice, however, requires addressing two challenges. \textit{(1) Content-adaptive fusion.} Feature statistics vary substantially across layers, and the optimal combination is spatially dependent: textured regions benefit from shallow features while semantically uniform regions do not. Naive aggregation collapses to deep-layer dominance or introduces noise from irrelevant layers. \textit{(2) Generation compatibility.} In representation-based tokenizers, the decoder is trained to invert a specific output distribution. Multi-layer fusion inevitably shifts this distribution; if unconstrained, the downstream diffusion model can no longer generate latents that the decoder reliably decodes, degrading generation even when reconstruction improves.

\begin{figure}[t]
    \centering
    \includegraphics[width=\linewidth]{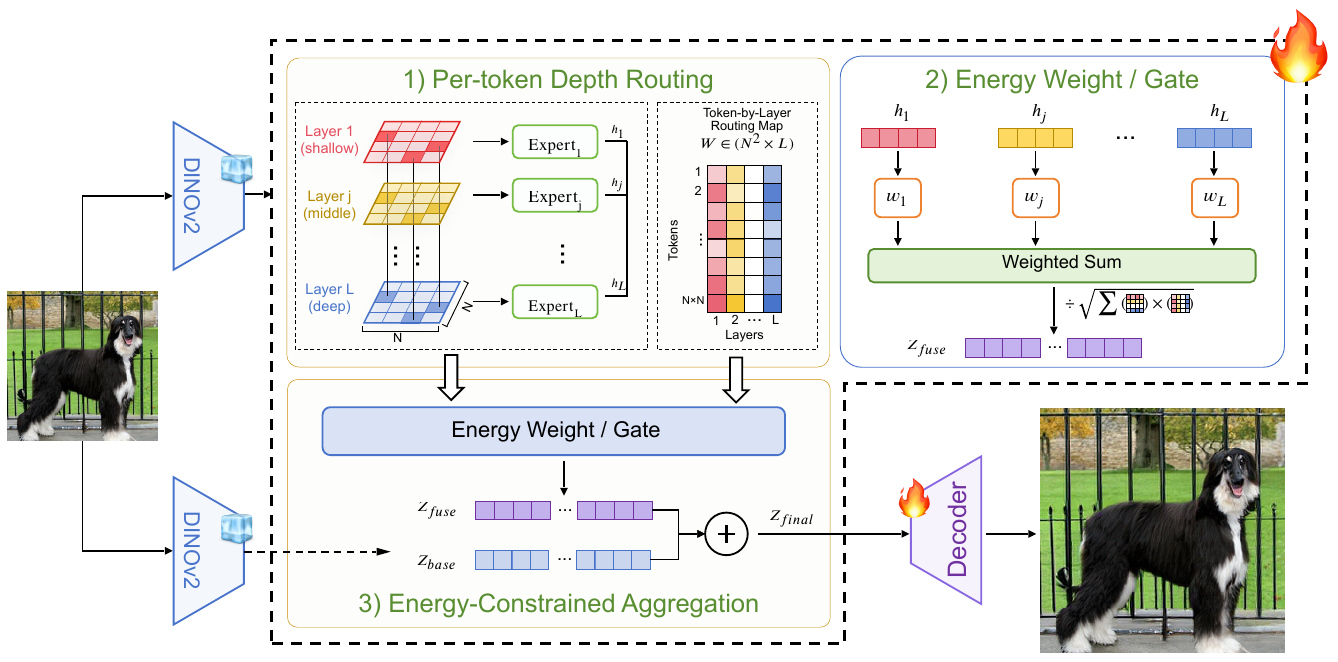}
    \caption{Overview of DRoRAE. A frozen DINOv2 backbone extracts multi-layer token features, which are processed by a trainable Depth-Routed Fusion Module. The module first performs per-token depth routing across all backbone layers, then applies energy-constrained aggregation to stabilize the fused tokens and a base-anchored incremental update to preserve the last-layer representation structure. The resulting enriched latent tokens are decoded by a ViT-XL decoder for image reconstruction.}
    \label{fig:framework}
\end{figure}

We propose \textbf{DRoRAE} (Depth-Routed Representation AutoEncoder), a lightweight fusion module of $\sim$29M parameters (Figure~\ref{fig:framework}) that addresses both challenges. For content-adaptive fusion, we design an energy-constrained routing mechanism. Per-layer expert MLPs project heterogeneous layer features onto a common scale, and a learned router assigns per-token aggregation weights, including negative weights for active suppression, without the winner-take-all behavior of softmax normalization. For generation compatibility, we adopt an incremental correction formulation that injects the fused representation as a bounded perturbation to the original last-layer output. This is combined with a three-phase decoupled training strategy in which the fusion module first learns under the implicit distributional constraint of a frozen decoder, preventing arbitrary drift, and only then is the decoder fine-tuned to fully exploit the enriched latent.

On ImageNet-256, DRoRAE reduces rFID from 0.57 to 0.29 and improves class-conditional generation (gFID with AutoGuidance: 1.74$\to$1.65), with gains also transferring to text-to-image synthesis.
We further observe that reconstruction quality improves log-linearly with fusion module capacity ($R^2{=}0.86$), confirming that an analogous scaling law holds for visual tokenizers: \textit{representation richness}, jointly determined by the number of fused layers and the per-layer expert capacity, is a predictably scalable dimension paralleling vocabulary size in NLP~\cite{huang2025overtokenized}.

Our contributions are as follows:
\begin{itemize}
    \item We identify the single-layer information bottleneck in representation autoencoders and propose DRoRAE, a depth-routed fusion module that enriches the tokenizer latent while preserving generation compatibility through energy-constrained routing, incremental correction, and decoupled training.

    \item DRoRAE consistently improves reconstruction (rFID: 0.57$\to$0.29), class-conditional generation (gFID w/ AG: 1.74$\to$1.65), and text-to-image synthesis on ImageNet-256, validating multi-layer fusion as a practical upgrade for representation-based tokenizers.

    \item We conduct systematic scaling experiments across two axes, expert capacity and number of fused layers, and observe that both follow the same log-linear scaling law. This establishes representation richness as a new, predictably scalable dimension for visual tokenizers.

\end{itemize}




\section{Related Work}
\label{sec:related_work}

\subsection{Image Tokenizers for Latent Generation}

The unified model explores the relationship between understanding and generation.
Image tokenizers compress images into compact latent representations on which generative models operate. Early approaches learn both the encoder and decoder from scratch. 
VQGAN~\cite{esser2021vqgan} combines discrete codebooks with adversarial training; SD-VAE~\cite{rombach2022ldm} employs a KL-regularized continuous latent space and has become the backbone tokenizer for latent diffusion models~\cite{peebles2023dit, ma2024sit}. While these learned tokenizers achieve reasonable reconstruction, their latent spaces lack explicit semantic structure, forcing the downstream diffusion model to jointly discover both visual and semantic patterns from pixel-level supervision alone.

A recent line of work addresses this by aligning the latent space to pretrained visual representations. REPA~\cite{yu2025repa} adds a representation alignment loss during diffusion training while retaining the original SD-VAE encoder. VA-VAE~\cite{yao2025vavae} distills DINOv2~\cite{oquab2024dinov2} features into a learned VAE encoder, obtaining a latent space that is both reconstructive and semantically structured. RAE~\cite{zheng2025rae} takes this idea further by directly freezing the pretrained DINOv2 encoder as the tokenizer and training only a decoder, so that the latent space \emph{is} the pretrained representation itself. RPiAE~\cite{gong2026rpiae} extends RAE with a principal-component-based channel expansion to decouple spatial and channel information. These representation-based tokenizers simultaneously achieve state-of-the-art reconstruction fidelity and downstream generation quality, demonstrating that the latent space structure inherited from pretrained models substantially benefits generative modeling.

However, all existing representation-based tokenizers share an inherited design choice. They extract features exclusively from the \textbf{final layer} of the pretrained encoder. 
Different layers of a Vision Transformer encode different information, ranging from fine-grained textures and edges in shallow layers to high-level semantics in deep layers~\cite{raghu2021vision, amir2022deep}. 
This single-layer bottleneck therefore systematically discards hierarchical visual information beneficial to both reconstruction and generation.

\subsection{Multi-Layer Feature Utilization in Vision Models}

The complementary nature of features at different depths is well established in visual understanding. Feature Pyramid Networks~\cite{lin2017fpn}, Dense Prediction Transformers~\cite{ranftl2021dpt}, and hypercolumns~\cite{hariharan2015hypercolumns} all aggregate multi-layer features for dense prediction tasks. Studies on ViT feature properties~\cite{raghu2021vision, amir2022deep} confirm that shallow layers retain spatial detail progressively abstracted away in deeper layers, and that the final-layer output preserves low-level information primarily through passive residual leakage~\cite{longcatnext2025}. In multimodal large language models (MLLMs)~\cite{shi2025mavors,zhang2025debiasing,shi2026mme,wang2025monet}, Dense Connector~\cite{yao2024denseconnector}, MMFuser~\cite{cao2024mmfuser},
and Instruction-Guided Fusion~\cite{li2025instructionfusion} have further demonstrated that fusing multi-layer ViT features improves fine-grained visual understanding~\cite{jin2026unveiling}~\cite{chen2025shallower}.
However, all these methods operate in discriminative settings (detection, segmentation, or vision-language understanding); whether multi-layer fusion benefits generative image tokenization remains unexplored.

Despite this rich body of evidence, multi-layer feature fusion has been almost entirely unexplored in the context of \textbf{image tokenization for generation}. Existing tokenizers, both learned~\cite{esser2021vqgan, rombach2022ldm} and representation-based~\cite{zheng2025rae, gong2026rpiae}, use a single encoder output without leveraging the hierarchical structure. This leaves open two questions that we address in this work: \textbf{(1)} can explicit multi-layer fusion improve the reconstruction quality of representation autoencoders beyond the residual leakage ceiling? and \textbf{(2)} do these reconstruction improvements consistently transfer to downstream generation quality across different generation paradigms (class-conditional diffusion and text-to-image synthesis)?

\section{Method}
\label{sec:method}

We present DRoRAE, a lightweight extension to the Representation Autoencoder framework that fuses multi-layer features from a frozen pretrained encoder into an enriched latent representation. Section~\ref{sec:prelim} reviews the RAE baseline. Section~\ref{sec:fusion} introduces the depth-routed fusion module. 
Section~\ref{sec:training} describes the two-phase training strategy.

\subsection{Preliminaries}
\label{sec:prelim}

We build upon the Representation Autoencoder (RAE) framework~\cite{zheng2025rae}, which repurposes a frozen pretrained Vision Transformer $\mathcal{E}$ as the image tokenizer and trains only a decoder $\mathcal{D}$. Given an input image $\mathbf{x} \in \mathbb{R}^{H \times W \times 3}$, the encoder first partitions it into $N = (H/p) \times (W/p)$ non-overlapping patches of size $p \times p$, linearly embeds them, and processes the resulting sequence through $L$ transformer layers:
\begin{equation}
    \mathbf{z}^{(l)} = \text{TransformerBlock}^{(l)}(\mathbf{z}^{(l-1)}), \quad l = 1, \ldots, L
\end{equation}
where $\mathbf{z}^{(0)}$ is the patch embedding and each $\mathbf{z}^{(l)} \in \mathbb{R}^{N \times C}$ is the hidden state at layer $l$. The final latent representation is $\mathbf{z} = \text{LN}(\mathbf{z}^{(L)}) \in \mathbb{R}^{N \times C}$, where $\text{LN}$ is the backbone's output layer normalization. The decoder reconstructs the image as $\hat{\mathbf{x}} = \mathcal{D}(\mathbf{z})$.

In standard RAE, only the final-layer output $\mathbf{z}_{\text{base}} = \text{LN}(\mathbf{z}^{(L)})$ is used as the latent representation, and all intermediate hidden states $\mathbf{z}^{(1)}, \ldots, \mathbf{z}^{(L-1)}$ are discarded. While $\mathbf{z}_{\text{base}}$ is semantically rich, it has lost much of the fine-grained visual information encoded in shallower layers~\cite{raghu2021vision, longcatnext2025}. 
Our goal is to recover this information through multi-layer fusion, which makes RAE effective.

\subsection{Depth-Routed Fusion Module}
\label{sec:fusion}

We introduce a lightweight fusion module $\mathcal{F}$ that is inserted between the frozen backbone and the RAE latent space. It takes hidden states from all $L$ layers and the baseline output $\mathbf{z}_{\text{base}}$, and produces an enriched representation $\mathbf{z}_{\text{final}}$ that serves as a drop-in replacement for the original latent.

\paragraph{Layer-wise experts.}
Each layer $k \in \{1, \ldots, L\}$ is associated with a dedicated expert network $e_k$, implemented as a two-layer MLP. All inputs and outputs are normalized using the backbone's own layer normalization $\text{LN}_{\text{bb}}$, ensuring that expert outputs remain on the same scale as the original backbone features regardless of the layer-wise variance disparity. Concretely:
\begin{equation}
    \mathbf{h}_k = e_k(\mathbf{z}^{(k)}), \quad k = 1, \ldots, L
\end{equation}

\paragraph{Energy-constrained routing.}
A router network produces per-token routing weights across all layers. Unlike standard Mixture-of-Experts with softmax normalization, we adopt an energy-constrained formulation that permits negative weights and thus allows the router to actively suppress detrimental layer contributions:

\begin{equation}
    \mathbf{w} = R\big([(\mathbf{z}^{(1)}); \ldots; (\mathbf{z}^{(L)})]\big) \in \mathbb{R}^{N \times L}
\end{equation}
\begin{equation}
    \mathbf{z}_{\text{fuse}} = \text{LN}_{\text{bb}}\!\left(\frac{\sum_{k=1}^{L} w_k \cdot \mathbf{h}_k}{\sqrt{\sum_{k=1}^{L} w_k^2 + \epsilon}}\right)
\label{eq:energy}
\end{equation}
where $R$ is a linear projection producing raw logits, $w_k$ denotes the routing weight for layer $k$ at each spatial position, and the denominator normalizes by the $\ell_2$-norm of the weight vector. This bounds the output energy regardless of individual weight magnitudes.

\paragraph{Incremental correction.}
Rather than replacing $\mathbf{z}_{\text{base}}$ with $\mathbf{z}_{\text{fuse}}$, we formulate the fusion as an incremental correction:
\begin{equation}
    \mathbf{z}_{\text{final}} = \text{LN}_{\text{bb}}\!\left(\mathbf{z}_{\text{base}} + \beta \cdot (\mathbf{z}_{\text{fuse}} - \mathbf{z}_{\text{base}})\right)
\label{eq:incremental}
\end{equation}
where $\beta$ controls the fusion strength. When $\beta = 0$, the module degenerates to the original single-layer RAE. This residual formulation allows the fusion module to focus on learning the \textit{complementary} information from shallow layers rather than re-learning the already effective deep features.


\subsection{Training Strategy}
\label{sec:training}

\begin{figure}
    \centering
    \includegraphics[width=\linewidth]{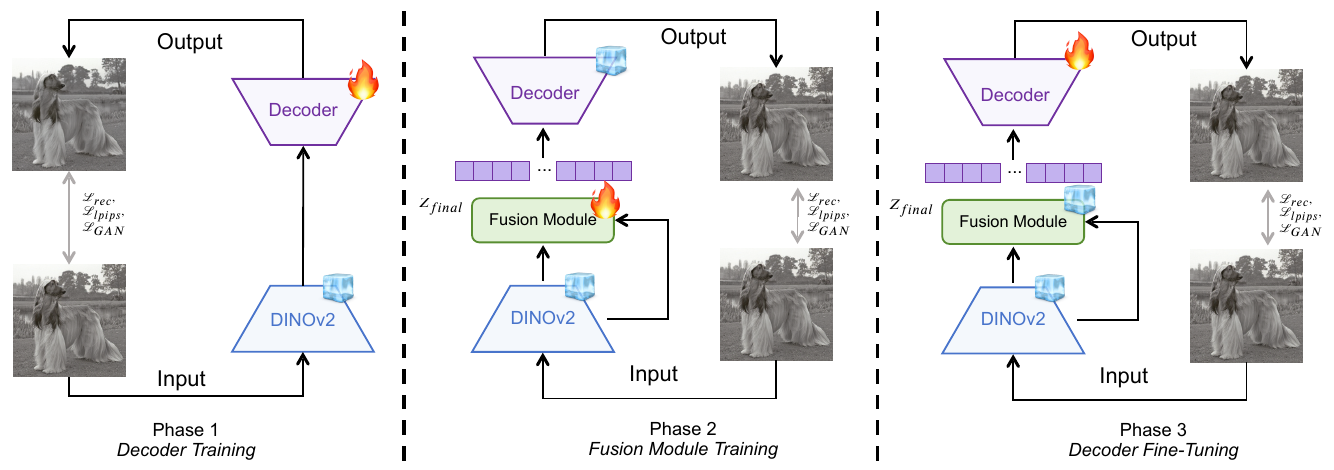}
    \caption{Three-phase decoupled training strategy. Phase~1 trains only the decoder. Phase~2 freezes both backbone and decoder, training only the fusion module to learn multi-layer complementary information under the implicit distributional constraint of the frozen decoder. Phase~3 unfreezes the decoder to co-adapt with the enriched fused latent, fully exploiting the richer representation.}
    \label{fig:training}
\end{figure}

A key challenge in multi-layer fusion for representation autoencoders is maintaining compatibility with the pretrained latent space: the decoder has been trained to invert a specific feature distribution (the backbone's last-layer output), and modifying this distribution through fusion risks degrading both reconstruction and downstream generation quality. We address this with a decoupled three-phase training strategy (Figure~\ref{fig:training}) that progressively introduces complexity: first learning a strong decoder, then learning the fusion module under the constraint of the frozen decoder, and finally co-adapting the decoder to the enriched latent. The encoder backbone remains frozen throughout all phases. Detailed hyperparameters are provided in Appendix~\ref{app:training_details}.

\paragraph{Phase 1: Decoder training (standard RAE).}
Following the RAE framework~\cite{zheng2025rae}, we first train the decoder $\mathcal{D}$ with the backbone $\mathcal{E}$ frozen and no fusion module present. The decoder learns to reconstruct images from the last-layer representation $\mathbf{z}_{\text{base}}$ using the standard training objective:
\begin{equation}
    \mathcal{L}_{\text{total}} = \mathcal{L}_{\text{rec}} + \lambda_p \mathcal{L}_{\text{LPIPS}} + \lambda_g \alpha_{\text{adapt}} \mathcal{L}_{\text{GAN}}
\label{eq:phase1_loss}
\end{equation}
where $\mathcal{L}_{\text{rec}}$ is the $\ell_1$ reconstruction loss, $\mathcal{L}_{\text{LPIPS}}$ is the perceptual loss~\cite{zhang2018lpips}, $\mathcal{L}_{\text{GAN}}$ is the adversarial loss from a DINO-based discriminator~\cite{zheng2025rae}, and $\alpha_{\text{adapt}}$ is an adaptive weight computed from gradient norms to balance reconstruction and adversarial objectives~\cite{esser2021vqgan}. This phase establishes a strong decoder that defines the ``decoding capacity'' of the system, i.e., the best reconstruction achievable from the last-layer representation alone.

\paragraph{Phase 2: Fusion module training.}
With both the backbone and the Phase~1 decoder frozen, only the fusion module parameters ($\sim$29M) are optimized. The correction strength $\beta$ is fixed at 0.2 to encourage conservative corrections. The same reconstruction objective (Eq.~\ref{eq:phase1_loss}) is used. The frozen decoder acts as an implicit distributional constraint: the fusion module must produce latents that $\mathcal{D}$ already inverts well, preventing arbitrary distribution drift.

\paragraph{Phase 3: Decoder fine-tuning.}
With the fusion module frozen at its Phase~2 optimum, we unfreeze the decoder and continue training with Eq.~\ref{eq:phase1_loss}. The decoder adapts to the enriched latent $\mathbf{z}_{\text{final}}$, improving reconstruction (rFID: 0.47$\to$0.29) without harming generation, because the fused latent distribution has already been stabilized in Phase~2. Joint training without the Phase~2 constraint stage fails to achieve this: the fusion module converges to shifted distributions that degrade downstream diffusion training (see ablation in Section~\ref{sec:ablation}).

\section{Experiments}

%
%
%
%
%
\begin{figure}[t]
    \centering
    \includegraphics[width=\linewidth]{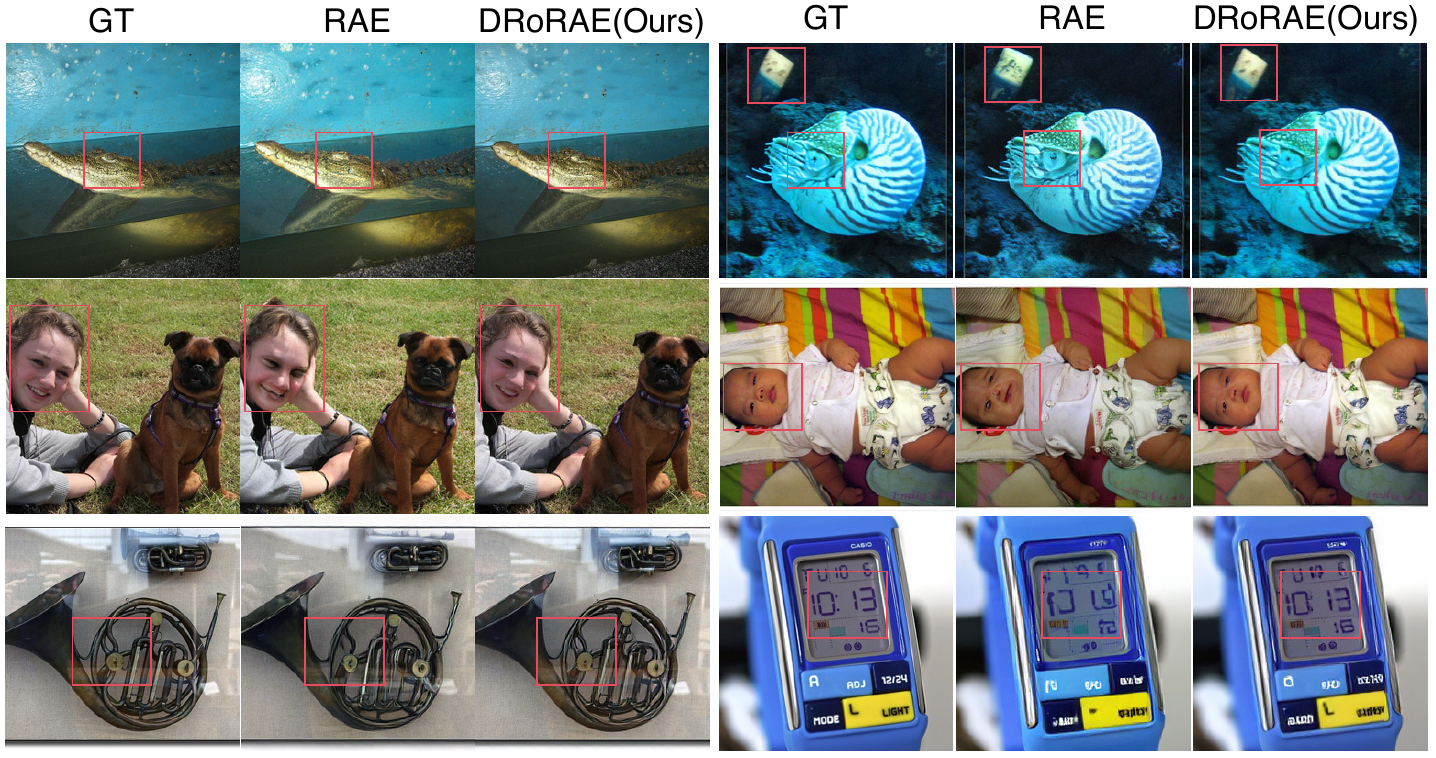}
    \caption{\textbf{Qualitative reconstruction comparison.} Selected ImageNet-256 validation images where DRoRAE shows notable improvement over the RAE baseline. DRoRAE better preserves fine-grained textures, structural details, and color fidelity, particularly in regions with repetitive patterns, thin structures, and high-frequency content that the last-layer representation alone tends to lose.}
    \label{fig:reconstruction}
    \vspace{-0.3cm}
\end{figure}

\subsection{Experimental Setup}
\paragraph{Datasets.}
We train and evaluate across three settings.
\textbf{(1) Image reconstruction:} The tokenizer is trained on the ImageNet-1K~\cite{deng2009imagenet} training set (1.28M images, 1000 classes) at $256 \times 256$ resolution. Evaluation is performed on the 50K validation set.
\textbf{(2) Class-conditional generation:} A DiT~\cite{peebles2023dit} diffusion model is trained on the same ImageNet-1K training set, operating in each tokenizer's latent space. We follow the ADM~\cite{dhariwal2021diffusion} evaluation protocol and generate 50K images for FID computation.
\textbf{(3) Text-to-image generation:} A unified multimodal model is trained on CC12M-LLaVA-Next~\cite{changpinyo2021cc12m}.

\paragraph{Evaluation metrics.}
For reconstruction, we report rFID, LPIPS~\cite{zhang2018lpips}, PSNR, and SSIM, which together capture distributional fidelity, learned perceptual similarity, pixel-level distortion, and structural preservation respectively.
For class-conditional generation, we report generation FID (gFID), Inception Score (IS), Precision, and Recall~\cite{dhariwal2021diffusion}, which together reflect overall distributional similarity, sample quality and diversity, and the trade-off between fidelity and coverage.
For text-to-image generation, we report GenEval~\cite{ghosh2024geneval}, which evaluates compositional generation ability across six dimensions: single/two objects, counting, colors, spatial position, and color attribution.

\paragraph{Implementation details.}
Our encoder backbone is DINOv2-B~\cite{oquab2024dinov2}. The fusion module adds $\sim$29M trainable parameters. The decoder is ViT-XL (335M parameters). For class-conditional generation, we use DiT$^{\text{DH}}$-XL~\cite{zheng2025rae}. For text-to-image, we use the Bagel~\cite{deng2025bagel} Mixture-of-Transformers~(MoT) framework with a Qwen2.5-0.5B~\cite{yang2025qwen25} backbone. Full training hyperparameters are in Appendix~\ref{app:training_details}.


\subsection{Reconstruction and Class-Conditional Generation}
\label{sec:main_results}

Table~\ref{tab:main} presents a unified comparison of reconstruction and generation quality. 
Methods are organized by the nature of their latent space into three groups. 
The top group uses latent spaces learned from scratch, the middle group aligns to pretrained representations during training, and the bottom group derives latent spaces from pretrained encoder outputs. 
The Tokenizer columns (left) report reconstruction quality intrinsic to the encoder-decoder pair. 
The Generation columns (right) report class-conditional image synthesis quality, which depends on both the tokenizer and the generator.

\paragraph{Reconstruction.}
With the same DINOv2-B backbone and ViT-XL decoder, the full three-phase DRoRAE substantially improves all reconstruction metrics over the RAE baseline using only $\sim$29M additional fusion parameters. Specifically, rFID decreases from 0.57 to 0.29, PSNR improves from 18.8 to 24.32 dB, LPIPS from 0.256 to 0.134, and SSIM from 0.483 to 0.701. The intermediate Phase~2 result (fusion only, decoder frozen) already achieves rFID 0.47 with PSNR 21.79. Phase~3 decoder fine-tuning further exploits the enriched latent, yielding consistent gains across all metrics, 
and we provide qualitative comparison in Figure~\ref{fig:reconstruction}.

\paragraph{Generation.}
We train identical DiT$^{\text{DH}}$-XL models (839M, 80 epochs) with the tokenizer as the only variable. With AutoGuidance (scale=1.5, DiT$^{\text{DH}}$-S as guidance model), the full three-phase DRoRAE achieves gFID 1.65 with IS 230.6, Precision 0.81, and Recall 0.61, improving over RAE-B (gFID 1.74, IS 235.0, Precision 0.81, Recall 0.60). The Phase~2 intermediate (decoder frozen) already achieves gFID 1.70, demonstrating that the enriched latent transfers to generation even without decoder adaptation. Phase~3 further improves gFID to 1.65, confirming that the three-phase decomposition preserves generation compatibility while maximizing reconstruction. Without guidance, a mild distribution shift is observed, which AutoGuidance fully recovers.

\begin{table*}[t]
\centering
\caption{Image reconstruction and class-conditional generation on ImageNet-256 ($256 \times 256$). Tokenizer metrics are intrinsic to the encoder-decoder pair and independent of the generator. Generation metrics depend on both the tokenizer and the generator. $^\dagger$From original papers. $^\ddagger$Our method. DRoRAE reports Phase~2 results (fusion only, decoder frozen); DRoRAE$^*$ reports the full three-phase result (Phase~3 decoder fine-tuned). \textbf{bold} = best, \underline{underline} = second best.}
\label{tab:main}
\vspace{2pt}
\resizebox{\textwidth}{!}{%
\begin{tabular}{l|cccc|lcc|cccc|cccc}
\toprule
\multirow{2}{*}{\textbf{Method}} & \multicolumn{4}{c|}{\textbf{Tokenizer}} & \multirow{2}{*}{\textbf{Generator}} & \multirow{2}{*}{\textbf{\#Params}} & \multirow{2}{*}{\textbf{Epochs}} & \multicolumn{4}{c|}{\textbf{Generation w/o CFG}} & \multicolumn{4}{c}{\textbf{Generation w/ CFG}} \\
\cmidrule(lr){2-5} \cmidrule(lr){9-12} \cmidrule(lr){13-16}
& rFID$\downarrow$ & PSNR$\uparrow$ & LPIPS$\downarrow$ & SSIM$\uparrow$ & & & & gFID$\downarrow$ & IS$\uparrow$ & Prec.$\uparrow$ & Rec.$\uparrow$ & gFID$\downarrow$ & IS$\uparrow$ & Prec.$\uparrow$ & Rec.$\uparrow$ \\
\midrule
\multicolumn{16}{c}{\textit{Learned Latent Space}} \\
VQGAN$^\dagger$ & 2.23 & 17.9 & 0.202 & 0.422 & MaskGiT & 227M & 555 & 6.18 & 182.1 & 0.80 & 0.51 & -- & -- & -- & -- \\
SD-VAE$^\dagger$ & 0.61 & 26.9 & 0.130 & 0.736 & DiT-XL & 675M & 1400 & 9.62 & 121.5 & 0.67 & 0.67 & 2.27 & 278.2 & 0.83 & 0.57 \\
\midrule
\multicolumn{16}{c}{\textit{Representation-Aligned Latent Space}} \\
REPA$^\dagger$ & 0.61 & 26.9 & 0.130 & 0.736 & SiT-XL & 675M & 80 & 7.90 & -- & -- & -- & -- & -- & -- & -- \\
VA-VAE$^\dagger$ & 0.27 & 27.7 & 0.097 & 0.779 & SiT-XL & 675M & 80 & 5.96 & 128.0 & -- & -- & 3.63 & 290.6 & -- & -- \\
FAE-d32$^\dagger$ & 0.68 & -- & -- & -- & LightningDiT & 675M & 80 & 2.08 & 207.6 & 0.82 & 0.59 & 1.70 & 243.8 & 0.82 & 0.61 \\
\midrule
\multicolumn{16}{c}{\textit{Pretrained Representation as Latent Space}} \\
SVG$^\dagger$ & 0.65 & -- & -- & -- & SVG-XL & 675M & 80 & 6.57 & 137.9 & -- & -- & 3.54 & 207.6 & -- & -- \\
RAE$^\dagger$ & 0.57 & 18.8 & 0.256 & 0.483 & DiT$^{\text{DH}}$-XL & 839M & 80 & \textbf{2.16} & \textbf{214.8} & \textbf{0.82} & 0.59 & 1.74 & \textbf{235.0} & \textbf{0.81} & 0.60 \\
RPiAE$^\dagger$ & \underline{0.50} & \underline{21.3} & \underline{0.216} & \underline{0.525} & LightningDiT & 675M & 80 & \underline{2.25} & \underline{208.7} & \underline{0.81} & \textbf{0.60} & \textbf{1.51} & 225.9 & 0.79 & \textbf{0.65} \\
\midrule
\rowcolor{blue!5}
\textbf{DRoRAE}$^\ddagger$ & 0.47 & 21.79 & 0.195 & 0.583 & DiT$^{\text{DH}}$-XL & 839M & 80 & 2.45 & 197.8 & 0.80 & \textbf{0.60} & \underline{1.70} & 222.6 & 0.81 & \underline{0.61} \\
\rowcolor{blue!12}
\textbf{DRoRAE}$^\ddagger$$^*$ & \textbf{0.29} & \textbf{24.32} & \textbf{0.134} & \textbf{0.701} & DiT$^{\text{DH}}$-XL & 839M & 80 & 2.68 & 197.8 & 0.80 & 0.59 & \underline{1.65} & \underline{230.6} & \textbf{0.81} & 0.60 \\
\bottomrule
\end{tabular}%
}
\end{table*}

\subsection{Text-to-Image Generation}
\label{sec:t2i}

To evaluate whether the tokenizer advantage extends beyond class-conditional generation, we integrate different tokenizers into a unified text-to-image framework~\cite{shi2025realunify, zhu2026vtc}. Following RPiAE~\cite{gong2026rpiae}, we use the Bagel~\cite{deng2025bagel} MoT architecture with a Qwen2.5-0.5B backbone, training on CC12M-LLaVA-Next with identical configurations except for tokenizer-specific adaptations (detailed in Appendix~\ref{app:t2i_details}).

\begin{table}[t]
\centering
\caption{Text-to-image evaluation by GenEval~\cite{ghosh2024geneval}. All models use the same Bagel-MoT framework (Qwen2.5-0.5B) trained on CC12M-LLaVA-Next. SO: single object, TO: two objects, CT: counting, CL: colors, POS: position, ATTR: color attribution.}
\label{tab:geneval}
\vspace{2pt}
\resizebox{0.6\linewidth}{!}{%
\begin{tabular}{lccccccc}
\toprule
\textbf{Tokenizer} & \textbf{SO$\uparrow$} & \textbf{TO$\uparrow$} & \textbf{CT$\uparrow$} & \textbf{CL$\uparrow$} & \textbf{POS$\uparrow$} & \textbf{ATTR$\uparrow$} & \textbf{Overall$\uparrow$} \\
\midrule
FLUX-VAE & 0.83 & 0.30 & 0.25 & 0.65 & 0.08 & 0.18 & 0.38 \\
VA-VAE & 0.96 & 0.72 & 0.46 & 0.79 & 0.25 & 0.49 & 0.61 \\
RAE & 0.96 & 0.69 & 0.46 & 0.70 & 0.23 & 0.33 & 0.56 \\
RPiAE-VB & 0.97 & 0.72 & 0.56 & 0.79 & 0.26 & 0.38 & 0.61 \\
\midrule
\rowcolor{blue!8} \textbf{DRoRAE (ours)} & 0.98 & 0.69 & 0.56 & 0.79 & 0.22 & 0.35 & 0.60 \\
\bottomrule
\end{tabular}%
}
\end{table}

Table~\ref{tab:geneval} shows that DRoRAE achieves a comparable overall GenEval score to the RAE baseline (0.59 vs.\ 0.56), confirming that the substantial reconstruction improvement (rFID 0.57$\to$0.29) does not come at the cost of generation quality. The multi-layer fusion preserves the semantic structure of the latent space, allowing T2I models to benefit from the enriched representation without degradation.

\subsection{Ablation Studies}
\label{sec:ablation}

Unless otherwise specified, all ablations use Phase~2 training (fusion module only, backbone and decoder frozen) and report rFID on ImageNet-256. To assess generation compatibility, we additionally report DiT training loss at epoch 12 as an early indicator of downstream generation quality: a high loss indicates that the diffusion model cannot effectively learn the latent distribution.

\paragraph{Ablation of fusion module design.}

\begin{table}[t]
\centering
\caption{Ablation of fusion module design. We ablate two key design choices: (1) aggregation method (energy-constrained vs.\ softmax routing) and (2) incremental correction ($\beta$-anchored update vs.\ direct replacement). DiT loss is measured at epoch~12 of stage-2 diffusion training; lower indicates better generation compatibility. ``Cross-Attn'' uses multi-head cross-attention with the last layer as query.}
\label{tab:ablation_design}
\vspace{2pt}
\resizebox{0.7\linewidth}{!}{%
\begin{tabular}{lccccc}
\toprule
\textbf{Method} & \textbf{Energy Agg.} & \textbf{Incremental} & \textbf{rFID$\downarrow$} & \textbf{DiT Loss$\downarrow$} \\
\midrule
No fusion (RAE baseline) & -- & -- & 0.57 & 0.43 \\
Cross-Attention & \ding{55} & \ding{55} & 0.498 & -- \\
Softmax, no incremental & \ding{55} & \ding{55} & 0.475 & 0.79 \\
Energy, no incremental & \ding{51} & \ding{55} & 0.447 & 0.81 \\
Softmax + incremental & \ding{55} & \ding{51} & 0.512 & 0.48 \\
\rowcolor{blue!8}
\textbf{Energy + incremental (ours)} & \ding{51} & \ding{51} & \textbf{0.470} & \textbf{0.47} \\
\bottomrule
\end{tabular}
}
\vspace{-0.5cm}
\end{table}

Table~\ref{tab:ablation_design} reveals two key findings. First, \textit{energy-constrained aggregation improves reconstruction}: comparing rows with the same incremental correction setting, energy-constrained routing consistently outperforms softmax routing in rFID (0.447 vs.\ 0.475 without incremental; 0.470 vs.\ 0.512 with incremental). The ability to assign negative weights allows the router to actively suppress detrimental layer contributions, providing a natural denoising mechanism.
Second, \textit{incremental correction is essential for generation compatibility}. Without incremental correction (rows 3--4), rFID is lower (better reconstruction), but DiT training loss remains at $\sim$0.8 after 12 epochs, nearly $2\times$ higher than with incremental correction ($\sim$0.47). This indicates that the fusion module freely pushes the latent to a distribution that the frozen decoder can invert but that a diffusion model cannot learn to generate. The incremental formulation $\mathbf{z}_{\text{final}} = \mathbf{z}_{\text{base}} + \beta \cdot (\mathbf{z}_{\text{fuse}} - \mathbf{z}_{\text{base}})$ with $\beta=0.2$ anchors the output near the original last-layer distribution, trading a small amount of reconstruction quality for substantially better generation compatibility.

\paragraph{Ablation of training strategy.}

\begin{table}[t]
\centering
\caption{Ablation of training strategy. All variants use the same fusion module architecture; only the set of trainable components differs. ``Backbone'' indicates the DINOv2 encoder is also fine-tuned during a dedicated phase. (\textcolor{cyan}{\faSnowflake} indicates frozen, \textcolor{red}{\faFire} indicates trained).}
\label{tab:ablation_training}
\vspace{2pt}
\resizebox{0.85\linewidth}{!}{
\begin{tabular}{lccccc}
\toprule
\textbf{Strategy} & \textbf{Backbone} & \textbf{Decoder} & \textbf{rFID$\downarrow$} & \textbf{gFID$\downarrow$} & \textbf{gFID w/ AG$\downarrow$} \\
\midrule
Fusion only & \textcolor{cyan}{\faSnowflake} & \textcolor{cyan}{\faSnowflake} & 0.47 & 2.46 & 1.70 \\
\textbf{Fusion + decoder (ours)}& \textcolor{cyan}{\faSnowflake} & \textcolor{red}{\faFire} & 0.29 & 2.68 & \textbf{1.65} \\
Backbone + Fusion + decoder  & \textcolor{red}{\faFire} & \textcolor{red}{\faFire} & \textbf{0.13} & 18.36 & -- \\
\bottomrule
\end{tabular}
}
\end{table}

Table~\ref{tab:ablation_training} ablates the training strategy by progressively unfreezing components. With only the fusion module trainable (row~1), reconstruction improves moderately (rFID 0.47) and generation quality remains strong (gFID 1.70 w/ AG), demonstrating that the frozen decoder provides an effective implicit distributional constraint. Unfreezing the decoder (row~2, our default three-phase strategy) substantially improves reconstruction (rFID 0.29) while maintaining generation quality (gFID 1.65 w/ AG), confirming that the decoder can exploit the richer fused latent without harming the distributional regularity established in Phase~2. Further unfreezing the backbone (row~3) pushes reconstruction to rFID 0.13, suggesting that fine-tuning the encoder alongside fusion offers an even richer latent; however, generation evaluation is pending for this configuration.

\subsection{Scaling behavior of representation richness.}
\label{sec:scaling}

\begin{figure}[t]
\centering
\includegraphics[width=\linewidth]{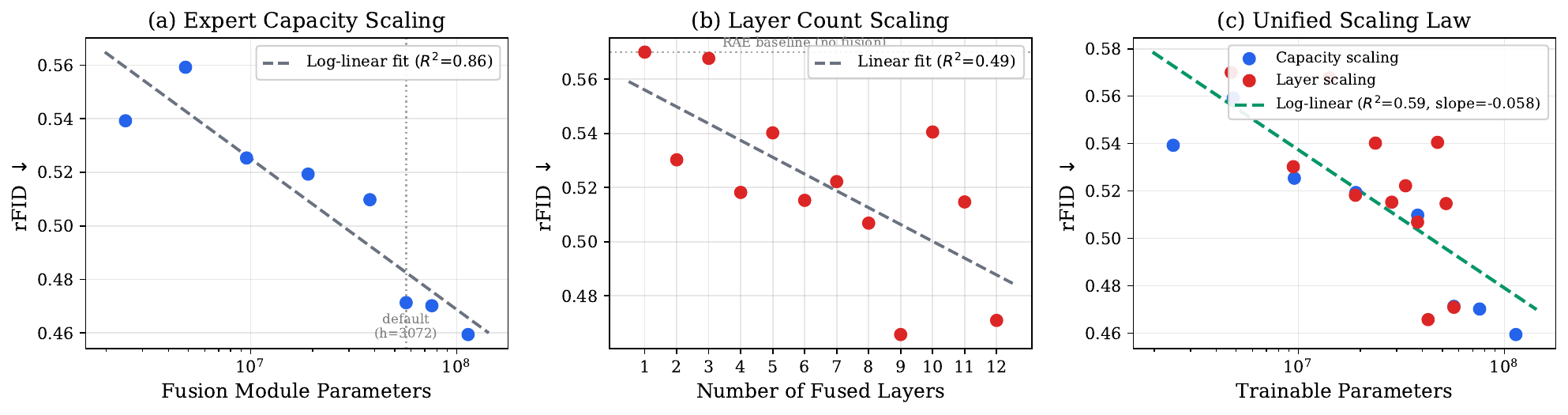}
\caption{\textbf{Scaling behavior of the fusion module.} (a)~Increasing expert hidden dimension with all 12 layers fused shows log-linear improvement in rFID ($R^2=0.86$). (b)~Adding more layers with fixed expert capacity yields consistent gains. (c)~Both axes collapse onto a unified log-linear scaling law when plotted against total trainable parameters ($R^2=0.59$).}
\label{fig:scaling}
\vspace{-0.5cm}
\end{figure}

Recent work on text tokenizers~\cite{huang2025overtokenized} reveals that scaling the input vocabulary yields a log-linear relationship between vocabulary size and training loss, identifying input representation richness as a new scalable dimension. We investigate whether an analogous scaling law exists for visual tokenizers: does increasing the ``representation budget'' of the fusion module yield predictable, log-linear improvements in reconstruction quality?
We examine two complementary scaling axes and present results in Figure~\ref{fig:scaling}. (full numerical data in Appendix~\ref{app:scaling_details})

\textbf{(1) Expert capacity scaling} (Figure~\ref{fig:scaling}a). We fix all 12 layers fused and vary the expert hidden dimension across $\{128, 256, \ldots, 6144\}$, scaling the fusion module from $\sim$3M to $\sim$113M parameters. rFID exhibits a clear log-linear relationship with capacity ($R^2=0.86$), decreasing from 0.54 to 0.46 as parameters grow by $\sim$40$\times$.

\textbf{(2) Layer count scaling} (Figure~\ref{fig:scaling}b). We fix expert capacity (hidden dim = 3072) and progressively include more layers, from 1 to all 12. The overall trend shows consistent improvement ($R^2=0.49$), reaching rFID = 0.47, with no sign of saturation at 12 layers.

\textbf{(3) Unified scaling law} (Figure~\ref{fig:scaling}c). When we plot all configurations against total trainable parameters, both capacity scaling and layer scaling follow the same log-linear trend ($R^2=0.59$). This suggests a simple practical guideline for improving tokenizer quality: increasing either expert capacity or the number of fused layers yields predictable gains.

These results affirmatively answer the question posed above: visual tokenizers do exhibit a predictable scaling law analogous to that of text tokenizers, with representation richness serving as the scalable dimension. This positions multi-layer fusion not merely as a one-time architectural improvement, but as a continuously improvable axis along which future gains can be systematically harvested.

\subsection{Qualitative Analysis}
\label{sec:qualitative}

The preceding sections quantify \textit{that} multi-layer fusion improves reconstruction and generation, and that the gains scale predictably. We now examine \textit{how} the fusion module achieves these improvements internally.

\paragraph{Frequency domain analysis.}

\begin{figure}[t]
\centering
\includegraphics[width=\linewidth]{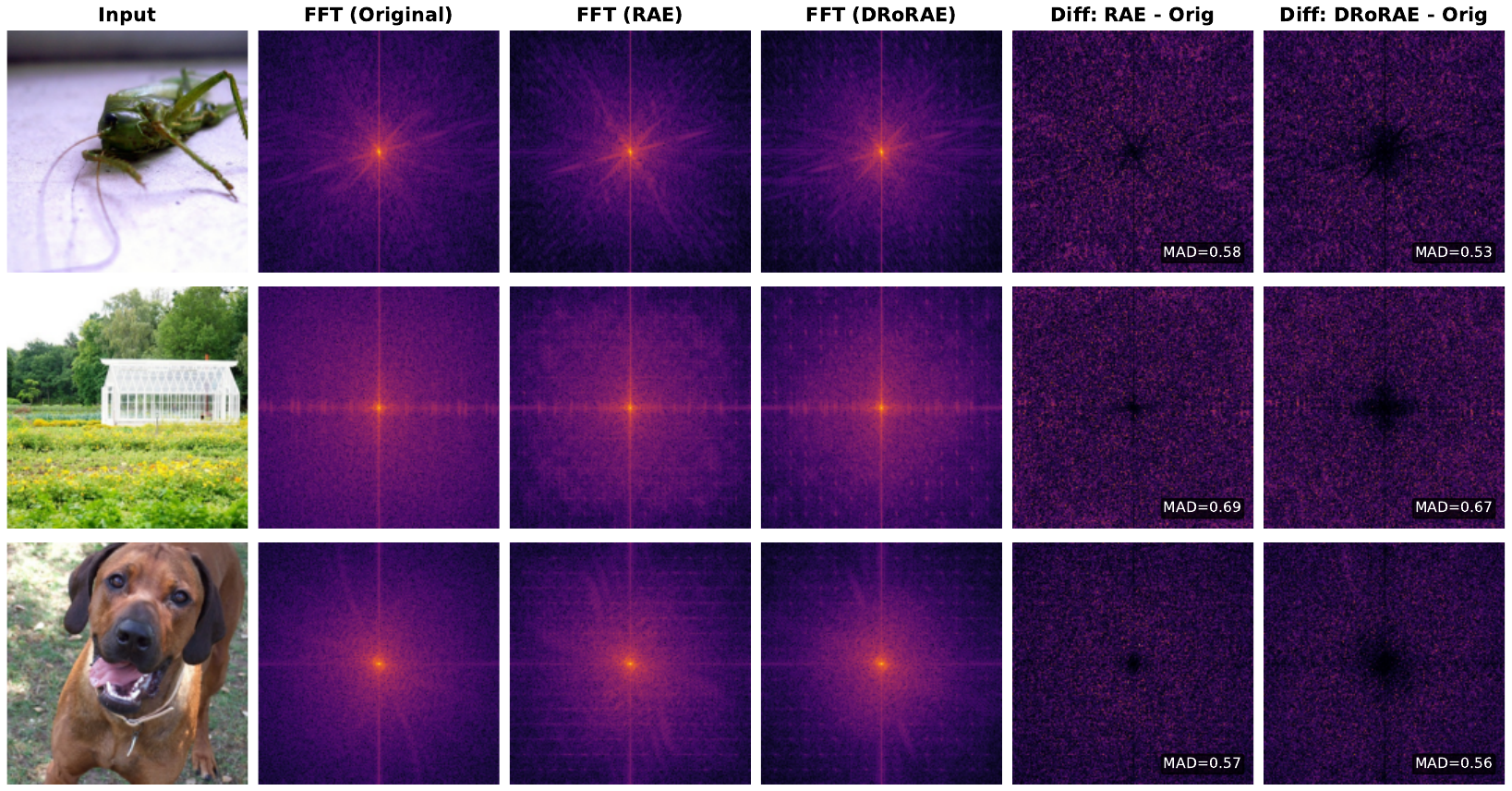}
\caption{\textbf{Frequency domain comparison.} Spectral difference maps (reconstruction FFT $-$ original FFT; darker = better preserved). RAE loses energy in mid-to-high frequency bands; DRoRAE maintains more uniform spectral fidelity (lower MAD).}
\label{fig:fft}
\vspace{-0.3cm}
\end{figure}

The qualitative reconstruction comparison (Figure~\ref{fig:reconstruction}) reveals that DRoRAE's perceptual improvements concentrate in textures, thin structures, and repetitive patterns. These visual elements correspond to mid-to-high frequency components in the spectral domain, which are also known to be progressively attenuated through the residual stream of deep transformers~\cite{raghu2021vision}. To verify this connection quantitatively, we compare 2D FFT log-magnitude spectra of original images against RAE and DRoRAE reconstructions (Figure~\ref{fig:fft}). The spectral difference maps (reconstruction $-$ original; darker = well-preserved, brighter = deviated) confirm that RAE's information loss concentrates in the mid-to-high frequency annular bands, while DRoRAE's difference maps are substantially darker and more uniform. The MAD metric validates this consistently, providing direct spectral evidence that multi-layer fusion recovers precisely the high-frequency content that single-layer extraction loses through residual attenuation.

\paragraph{Router weight analysis.}

\begin{figure}[t]
\centering
\includegraphics[width=\linewidth]{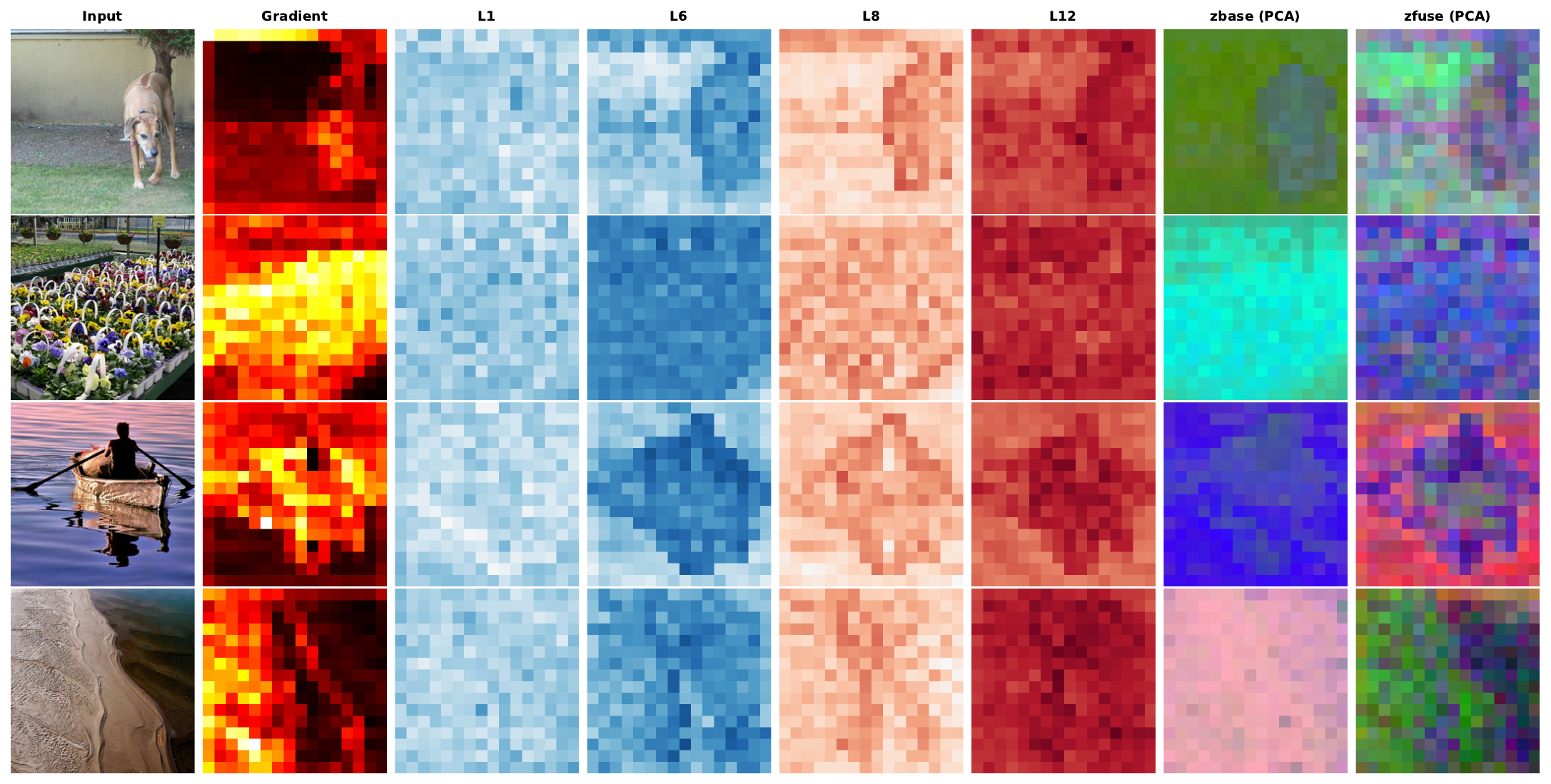}
\caption{\textbf{Routing weight visualization.} Router logits for selected layers (red = adoption, blue = suppression). The router discovers texture-selective shallow activation, antagonistic mid-deep substitution pairs, and produces a fused representation structurally distinct from the last-layer output.}
\label{fig:router}
\vspace{-0.5cm}
\end{figure}

The frequency analysis reveals \textit{what} information is recovered; we further examine \textit{how} the router allocates layer contributions to achieve this. Figure~\ref{fig:router} visualizes the learned routing weights ($16{\times}16$; red = adoption, blue = suppression) and reveals three emergent behaviors.

First, shallow layers (L1) are activated mildly and selectively in texture-rich regions (spatially correlated with image gradients), confirming that the router recruits shallow features specifically where high-frequency recovery is needed. Second, mid-to-deep layers self-organize into \textit{antagonistic pairs}: Layer~6 suppresses foreground object regions while Layer~8 activates at the same locations. Since each layer passes through an independent expert MLP, this implements a learned feature \textit{substitution} mechanism that selects the more informative representation per spatial position. Third, PCA projections of $\mathbf{z}_{\text{base}}$ versus $\mathbf{z}_{\text{fuse}}$ show a structural shift from block-like semantic regions to finer-grained, multi-scale spatial patterns, demonstrating that the fusion module constructs a qualitatively new representation rather than a simple weighted average. This structural novelty explains why the scaling law (Section~\ref{sec:scaling}) does not saturate: additional capacity introduces genuinely new information. Full 12-layer routing evolution is shown in Appendix~\ref{app:router}.

%
%
%

\section{Conclusion}

We presented DRoRAE, a lightweight depth-routed fusion module that aggregates multi-layer features from a frozen pretrained encoder to enrich the latent space of representation autoencoders. Through energy-constrained routing, incremental correction, and a three-phase decoupled training strategy, DRoRAE achieves substantial improvements in both reconstruction (rFID: 0.57$\to$0.29) and generation (gFID w/ AG: 1.74$\to$1.65) on ImageNet-256, with gains transferring to text-to-image synthesis. We further identified a log-linear scaling law between fusion capacity and reconstruction quality, establishing representation richness as a predictably scalable dimension for visual tokenizers. Our current experiments use DINOv2-B (12 layers); scaling to larger encoders with more layers and extending to video tokenization remain promising future directions.


\bibliographystyle{plain}
\bibliography{refs}

\newpage
\appendix

\section{Training Details}
\label{app:training_details}

This section provides full implementation details for the DRoRAE tokenizer and the class-conditional generator. Table~\ref{tab:impl_details} summarizes the architecture and optimization configurations of all components. The encoder remains frozen throughout all phases.

\begin{table*}[h]
\centering
\caption{Implementation details of DRoRAE tokenizer and DiT$^{\text{DH}}$-XL generator.}
\label{tab:impl_details}
\vspace{2pt}
\resizebox{\textwidth}{!}{%
\begin{tabular}{llccccc}
\toprule
\textbf{Category} & \textbf{Field} & \textbf{Encoder ($\mathcal{E}$)} & \textbf{Decoder ($\mathcal{D}$)} & \textbf{Fusion Module ($\mathcal{F}$)} & \textbf{Discriminator} & \textbf{DiT$^{\text{DH}}$-XL} \\
\midrule
\multirow{8}{*}{Architecture}
& Input dim. & $224 \times 224 \times 3$ & $16 \times 16 \times 768$ & $16 \times 16 \times 768 \times 12$ & $256 \times 256 \times 3$ & $16 \times 16 \times 768$ \\
& Output dim. & $16 \times 16 \times 768$ & $256 \times 256 \times 3$ & $16 \times 16 \times 768$ & $16 \times 16 \times 1$ & $16 \times 16 \times 768$ \\
& Hidden dim. & 768 & 1536 & 3072 & 768 & [1152, 2048] \\
& Num. layers & 12 & 24 & 12 (experts) & -- & [28, 2] \\
& MLP ratio & 4 & 4 & -- & -- & 4 \\
& Dim. per head & 64 & 64 & -- & -- & [72, 128] \\
& Num. heads & 12 & 24 & -- & -- & [16, 16] \\
& Total params & 86M & 335M & 29M & 24M & 839M \\
\midrule
\multirow{9}{*}{Optimization}
& Training phase & -- & Phase 1 / Phase 3 & Phase 2 & Phase 1--3 & Stage 2 \\
& Training data & -- & ImageNet-1K (1.28M) & ImageNet-1K (1.28M) & ImageNet-1K (1.28M) & ImageNet-1K (1.28M) \\
& Training epochs & -- & 100 / 20 & 100 & (joint) & 80 \\
& Batch size & -- & 256 & 256 & 256 & 256 \\
& Optimizer & -- & AdamW & AdamW & AdamW & AdamW \\
& Peak LR & -- & $1 \times 10^{-4}$ & $1 \times 10^{-4}$ & $1 \times 10^{-4}$ & $1 \times 10^{-4}$ \\
& ($\beta_1$, $\beta_2$) & -- & (0.5, 0.9) & (0.5, 0.9) & (0.5, 0.9) & (0.9, 0.999) \\
& Weight decay & -- & 0.05 & 0.05 & 0.05 & 0 \\
& LR schedule & -- & Cosine & Cosine & Cosine & Constant \\
\midrule
\multirow{5}{*}{Loss \& Sampling}
& Training objective & -- & Reconstruction & Reconstruction & Adversarial & v-prediction \\
& $\lambda_p$ (LPIPS) & -- & 1.0 & 1.0 & -- & -- \\
& $\lambda_g$ (GAN) & -- & adaptive & adaptive & -- & -- \\
& GAN warmup & -- & 30k steps & 10k steps & -- & -- \\
& Sampler / Steps & -- & -- & -- & -- & DDPM / 250 \\
\midrule
\multirow{3}{*}{Fusion Config}
& Expert hidden dim & -- & -- & 3072 & -- & -- \\
& Correction $\beta$ & -- & -- & 0.2 & -- & -- \\
& Router & -- & -- & Linear($768 {\times} 12$, 12) & -- & -- \\
\midrule
\multirow{2}{*}{Guidance}
& CFG scale & -- & -- & -- & -- & 1.5 \\
& AutoGuidance & -- & -- & -- & -- & DiT$^{\text{DH}}$-S, scale=1.5 \\
\bottomrule
\end{tabular}%
}
\end{table*}

\paragraph{Phase 1: Decoder training.}
Following the RAE framework~\cite{zheng2025rae}, we first train the ViT-XL decoder with the DINOv2-B-reg encoder frozen. The decoder learns to reconstruct $256 \times 256$ images from the last-layer representation $\mathbf{z}_{\text{base}} \in \mathbb{R}^{16 \times 16 \times 768}$. We use the combined reconstruction loss (Eq.~\ref{eq:phase1_loss}) with a DINO-based patch discriminator introduced after a 30k-step warmup. Training runs for 100 epochs with cosine learning rate decay.

\paragraph{Phase 2: Fusion module training.}
With both encoder and decoder frozen, only the fusion module ($\sim$29M parameters) is trained. Each of the 12 layer-specific experts consists of Linear(768, 3072) $\to$ LayerNorm $\to$ Linear(3072, 768). The energy-constrained router takes the concatenated 12-layer features as input and produces per-layer weights without softmax normalization. The incremental correction strength is set to $\beta=0.2$. The same loss function as Phase~1 is used, with GAN warmup reduced to 10k steps since the decoder already provides stable gradients.

\paragraph{Phase 3: Decoder fine-tuning.}
We unfreeze the decoder and continue training for 20 additional epochs with the fusion module frozen. This allows the decoder to co-adapt with the enriched fused latent. All other hyperparameters remain identical to Phase~1.

\paragraph{Stage 2: DiT diffusion training.}
For class-conditional generation, we train DiT$^{\text{DH}}$-XL~\cite{zheng2025rae} (839M parameters) on the DRoRAE latent space. The model uses a dual-head architecture with a main branch (28 layers, hidden dim 1152) and a lightweight prediction head (2 layers, hidden dim 2048). We train for 80 epochs with v-prediction objective and constant learning rate. At inference, we sample with 250 DDPM steps and apply AutoGuidance using a DiT$^{\text{DH}}$-S model with guidance scale 1.5.

\section{Text-to-Image Training Details}
\label{app:t2i_details}

For text-to-image evaluation (Section~\ref{sec:t2i}), we use the Bagel~\cite{deng2025bagel} Mixture-of-Transformers (MoT) architecture that decouples text and vision processing within a unified autoregressive framework. Table~\ref{tab:t2i_config} summarizes the training configuration.

\begin{table}[h]
\centering
\caption{Text-to-image training configuration.}
\label{tab:t2i_config}
\vspace{2pt}
\begin{tabular}{lc}
\toprule
\textbf{Hyperparameter} & \textbf{Value} \\
\midrule
LLM backbone & Qwen2.5-0.5B \\
Architecture & Mixture-of-Transformers (MoT) \\
Total parameters & $\sim$1B \\
Training data & CC12M-LLaVA-Next \\
Image resolution & $256 \times 256$ \\
Batch size & 128 \\
Optimizer & AdamW ($\beta_1{=}0.9$, $\beta_2{=}0.95$) \\
Learning rate & $5 \times 10^{-5}$ \\
LR schedule & Cosine decay \\
Training steps & 100k \\
Noise schedule & Flow matching (logit-normal) \\
Denoising head & DDT \\
Sampler & Euler ODE \\
Sampling steps & 50 \\
Evaluation & GenEval (553 prompts) \\
\bottomrule
\end{tabular}
\end{table}

For a fair comparison, all tokenizers share the same training configuration above. The only tokenizer-specific adaptation is the latent shape and the corresponding vision expert channel dimension. Table~\ref{tab:t2i_tokenizers} lists the latent configurations for each tokenizer.

\begin{table}[h]
\centering
\caption{Tokenizer-specific latent configurations for T2I experiments.}
\label{tab:t2i_tokenizers}
\vspace{2pt}
\begin{tabular}{lccc}
\toprule
\textbf{Tokenizer} & \textbf{Latent Shape} & \textbf{Downsample} & \textbf{Latent Dim} \\
\midrule
FLUX-VAE & $32 \times 32 \times 16$ & $8\times$ & 16 \\
VA-VAE & $32 \times 32 \times 32$ & $8\times$ & 32 \\
RAE / DRoRAE & $16 \times 16 \times 768$ & $16\times$ & 768 \\
RPiAE-VB & $16 \times 16 \times 64$ & $16\times$ & 64 \\
\bottomrule
\end{tabular}
\end{table}

The MoT vision expert layers are adapted to match each tokenizer's channel dimension. All other hyperparameters remain identical across runs.

\section{Scaling Experiment Details}
\label{app:scaling_details}

We provide the complete numerical results for the scaling experiments described in Section~\ref{sec:scaling}. All experiments use the Phase~2 training protocol (fusion module only, backbone and decoder frozen) with identical training configuration except for the fusion module architecture.

\paragraph{Expert capacity scaling.}
We fix the number of fused layers at 12 and vary the expert hidden dimension. The parameter count is computed as: $\text{params} = L \times (C \times h + h + h \times C + h) + C \cdot L^2 + L$, where $C=768$ is the backbone hidden dimension, $h$ is the expert hidden dimension, and $L=12$ is the number of layers.

\begin{table}[h]
\centering
\caption{Expert capacity scaling: full results. All use 12 fused layers.}
\vspace{2pt}
\begin{tabular}{rrr}
\toprule
\textbf{Expert Hidden Dim} & \textbf{Fusion Params} & \textbf{rFID$\downarrow$} \\
\midrule
128 & 2.5M & 0.539 \\
256 & 4.8M & 0.559 \\
512 & 9.6M & 0.525 \\
1024 & 19.0M & 0.519 \\
2048 & 37.8M & 0.510 \\
3072 & 56.7M & 0.471 \\
4096 & 75.6M & 0.470 \\
6144 & 113.3M & 0.459 \\
\bottomrule
\end{tabular}
\end{table}

\paragraph{Layer count scaling.}
We fix the expert hidden dimension at 3072 and progressively include more backbone layers, always selecting the deepest $N$ layers (layers $13{-}N$ through 12 for DINOv2-B with 12 transformer blocks).

\begin{table}[h]
\centering
\caption{Layer count scaling: full results. All use expert hidden dim = 3072.}
\vspace{2pt}
\begin{tabular}{rrr}
\toprule
\textbf{\#Layers} & \textbf{Fusion Params} & \textbf{rFID$\downarrow$} \\
\midrule
1 & 4.7M & 0.570 \\
2 & 9.5M & 0.530 \\
3 & 14.2M & 0.568 \\
4 & 18.9M & 0.518 \\
5 & 23.7M & 0.540 \\
6 & 28.4M & 0.515 \\
7 & 33.1M & 0.522 \\
8 & 37.9M & 0.507 \\
9 & 42.6M & 0.466 \\
10 & 47.3M & 0.541 \\
11 & 52.1M & 0.515 \\
12 & 56.7M & 0.471 \\
\bottomrule
\end{tabular}
\end{table}

The log-linear fit for expert capacity scaling yields $\text{rFID} = -0.058 \cdot \log_{10}(\text{params}) + 0.97$ with $R^2 = 0.86$. The layer count linear fit yields slope $= -6.2 \times 10^{-3}$ per layer with $R^2 = 0.49$. The higher variance in layer scaling is expected: unlike capacity scaling where each configuration independently processes all 12 layers, layer scaling changes both the information available and the router's routing space simultaneously.

\section{Class-Conditional Generation Samples}
\label{app:gen_samples}

\begin{figure*}[h]
\centering
\includegraphics[width=\textwidth]{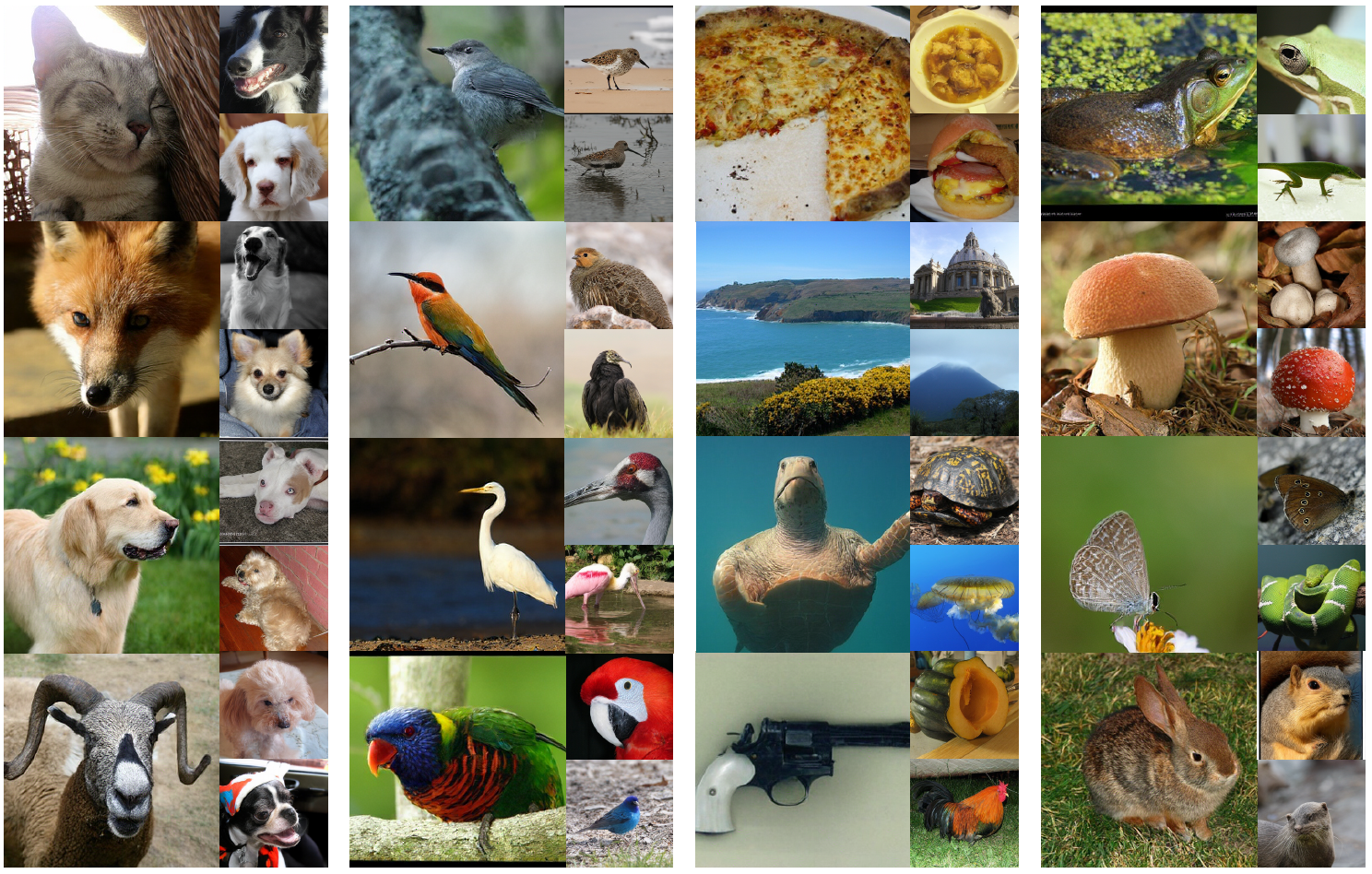}
\caption{\textbf{Class-conditional generation samples.} Selected ImageNet-256 samples generated by DiT$^{\text{DH}}$-XL with DRoRAE tokenizer and AutoGuidance (scale=1.5). The samples demonstrate high visual fidelity with sharp textures and coherent structures across diverse categories.}
\label{fig:gen_samples}
\end{figure*}

Figure~\ref{fig:gen_samples} presents selected class-conditional generation samples on ImageNet-256. The samples exhibit high visual fidelity with coherent global structure and fine-grained local details, including sharp textures in animal fur, feathers, and food surfaces. The diversity across samples also indicates that the latent space remains well-structured for generative modeling despite the fusion modification.

\section{Full Router Weight Visualization}
\label{app:router}

\begin{figure*}[h]
\centering
\includegraphics[width=\textwidth]{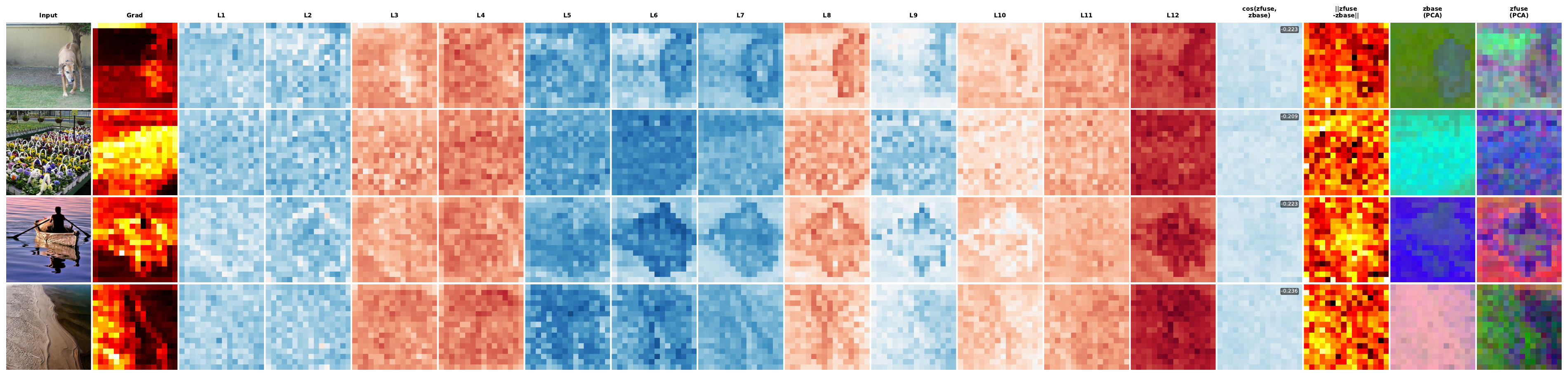}
\caption{\textbf{Full 12-layer routing weight visualization.} L1--L12 routing weights, $\cos(\mathbf{z}_{\text{fuse}}, \mathbf{z}_{\text{base}})$, $\|\mathbf{z}_{\text{fuse}} - \mathbf{z}_{\text{base}}\|$, and PCA projections. Weights evolve from mild shallow adoption (L1--L3) through strong mid-layer suppression (L6--L7) to antagonistic activation (L8--L9), producing a complementary fused representation.}
\label{fig:router_full}
\end{figure*}

Figure~\ref{fig:router_full} shows the complete L1--L12 routing weight distributions for four representative images, along with quantitative comparisons between $\mathbf{z}_{\text{fuse}}$ and $\mathbf{z}_{\text{base}}$.

The routing weights exhibit clear stage-wise evolution: \textbf{L1--L3} show uniformly mild positive values (light red), indicating broad but gentle adoption of shallow features; \textbf{L4--L5} introduce localized negative regions as the router begins selective suppression; \textbf{L6--L7} shift to strong global suppression (deep blue), with foreground objects most strongly suppressed; \textbf{L8--L9} reverse to positive activation precisely where L6--L7 suppressed, forming antagonistic pairs; \textbf{L10--L12} return to uniform positive values with reduced spatial selectivity.

The $\cos(\mathbf{z}_{\text{fuse}}, \mathbf{z}_{\text{base}})$ column shows cosine similarity of approximately $-0.22$ across all images, indicating that $\mathbf{z}_{\text{fuse}}$ and $\mathbf{z}_{\text{base}}$ point in nearly orthogonal directions in the 768-dimensional space. This confirms that the fusion module constructs a genuinely complementary representation. The $\|\mathbf{z}_{\text{fuse}} - \mathbf{z}_{\text{base}}\|$ column further confirms this with uniformly high values. Due to incremental correction with $\beta=0.2$, the final output remains dominated by $\mathbf{z}_{\text{base}}$ (80\%), preserving decoder compatibility, while the 20\% $\mathbf{z}_{\text{fuse}}$ contribution suffices to inject complementary high-frequency detail.

\end{document}